\documentclass[conference]{IEEEtran}
\IEEEoverridecommandlockouts
\usepackage{cite}
\usepackage{amsmath,amssymb,amsfonts}
\usepackage{algorithmic}
\usepackage{graphicx}
\usepackage{textcomp}
\usepackage{xcolor}
\usepackage{lipsum}
\usepackage{wrapfig}
\usepackage{multirow}
\usepackage{tabularx}
\usepackage{pifont}

\def\BibTeX{{\rm B\kern-.05em{\sc i\kern-.025em b}\kern-.08em
    T\kern-.1667em\lower.7ex\hbox{E}\kern-.125emX}}
    
\begin{document}

\title{Transfer learning with class-weighted and focal loss function 
for automatic skin cancer classification
\\}

\author{\IEEEauthorblockN{Duyen N.T. Le\textsuperscript{1}, Hieu X. Le\textsuperscript{1}, Lua T. Ngo\textsuperscript{1}, and Hoan T. Ngo\textsuperscript{1*}}
\IEEEauthorblockA{\textit{\textsuperscript{1}School of Biomedical Engineering, International University, Vietnam
National University, 
Ho Chi Minh
City, Vietnam
}\\
\textit{\textsuperscript{*}Corresponding author: nthoan@hcmiu.edu.vn}
} 
}

\maketitle

\begin{abstract}
Skin cancer is by far in top-3 of the world’s most common cancer. Among different skin cancer types, melanoma is particularly dangerous because of its ability to metastasize. Early detection is the key to success in skin cancer treatment. However, skin cancer diagnosis is still a challenge, even for experienced dermatologists, due to strong resemblances between benign and malignant lesions. To aid dermatologists in skin cancer diagnosis, we developed a deep learning system that can effectively and automatically classify skin lesions into one of the seven classes: (1) Actinic Keratoses, (2) Basal Cell Carcinoma, (3) Benign Keratosis, (4) Dermatofibroma, (5) Melanocytic nevi, (6) Melanoma, (7) Vascular Skin Lesion. The HAM10000 dataset was used to train the system. An end-to-end deep learning process, transfer learning technique, utilizing multiple pre-trained models, combining with class-weighted and focal loss were applied for the classification process. The result was that our ensemble of modified ResNet50 models can classify skin lesions into one of the seven classes with top-1, top-2 and top-3 accuracy 93\%, 97\% and 99\%, respectively.  This deep learning system can potentially be integrated into computer-aided diagnosis systems that support dermatologists in skin cancer diagnosis.
\end{abstract}

\begin{IEEEkeywords}
Deep learning, skin lesions classification, HAM10000, ensemble model
\end{IEEEkeywords}

\section{Introduction}
Skin cancer incidence has climbed gradually over the past 30 years and it is forecasted to increase in near future \cite{b1}\cite{b2}. This malignancy is classified into two sub-types: non-melanoma and melanoma on the ground that melanoma is the most aggressive form due to its tendency to rapidly metastasize. According to WHO, between 2 to 3 million non-melanoma skin cancer cases and about 132,000 melanoma skin cancer cases will occur globally each year and one out of three identified cancer cases is skin cancer \cite{b3}. In general, skin cancer incidence rate (both non-melanoma and melanoma) was at the 3rd highest amongst other cancers in 2018 with  1,329,781 new cases \cite{b4}. Although melanoma is not as common as other skin cancers, about 1 percent of skin cancer cases, it accounts for majority of deaths related to skin cancer \cite{b1}. In U.S alone, melanoma causes over 10,000 deaths every year \cite{b5}. Skin cancer deaths have been increasing over the past decades \cite{b6} and are projected to climb in the next four decades \cite{b7}. Thus, it could be a massive burden for society and healthcare systems \cite{b8}. 

On the other hand, melanoma is curable with an estimated 5-year survival rate approximately 95 percent if it is detected at early stages \cite{b3}. Note that this survival rate is different amongst ethnicities. For people of color, because of high amount of melamine in skin tissue, their risk of having skin cancer is low. However, also due to this fact, skin cancer is often in late stage when diagnosed in people of color. Consequently, their 5-year survival rate reduce to 70\% \cite{b9}.

Malignant skin lesions, which bear a strong resemblance to benign skin lesions, are considerably difficult to identify. Dermatologists with naked eyes can diagnose skin cancer with an accuracy of ~60\%. Equipped with dermatoscope and been well-trained, they can achieve better accuracy of 75\%-84\% \cite{b10}\cite{b11}\cite{b12}. However, it strongly relies on dermatologists’ experience. Over the past decade, application of artificial intelligence (AI), particularly deep learning, for skin cancer, particularly melanoma, diagnosis has drawn a lot of interest. With the introduction of various methodologies \cite{b13}, deep learning’s accuracy in skin cancer diagnosis has rapidly raised and, in some cases, outperformed individual dermatologists. It is believed that the development of AI based tools integrated into mobile devices can extend the reach of medical care in such a way that patients can access to dermatological expertise regardless of where they live \cite{b14}.   

This paper proposes a deep convolutional neural network (CNN) with high accuracy for classifying skin lesions into seven categories. Transfer learning technique was applied by employing pre-trained ResNet50, VGG16 and MobileNet models in combination with focal loss and class weighted. CNN-based deep learning provides an end-to-end approach, which achieves high accuracy without the need of feature engineering.
 
\section{Related works}
Since the 1990s, there have been many efforts in using computer algorithms to overcome challenges in skin cancer diagnosis. Several studies began when Binder et al. developed an ANN for classifying benign and cancerous lesions by dermatoscopic images with 100 images for training and testing \cite{b15}. His ANN model achieved 88\% specificity compared with 90\% in human diagnosis. Consequently, a vast array of methods has been conducted aiming to diagnose melanoma via dermoscopy or clinical images. Those approaches ranging from computational analysis as fuzzy logic to machine learning algorithms as support vector machine \cite{b16}, K-nearest neighbor \cite{b17}, Naïve Bayesian \cite{b18},… However, these studies involved in heavy workloads for image preprocessing, feature extraction, and feature selection.

Recently, with high accuracy and an end-to-end approach without feature engineering, convolutional neural networks became a preferable solution for skin lesions classification \cite{b19}. Transfer learning is a technique that repurposes the existing models which were pre-trained on large datasets for automatic extraction of features from new datasets. The most common pre-trained models applied in skin lesions classification are ResNet50, VGG, and MobileNet. Some hybrid schemes combining deep learning and other machine learning techniques for skin lesions classification were also investigated and can achieve 84\%  accuracy \cite{b17}. They used deep learning model as an auto feature extraction and feed those results into SVM, decision tree,… to classify skin lesions. However, performance comparison of CNN models and other machine learning methods in \cite{b20} suggested better results come from state-of-the-art deep learning models. In 2017, Esteva et al used Inception v3 pretrained model from Google and achieved 72.1\% classification accuracy after training on 129,450 clinical images for 2032 skin diseases, whose competence is comparable to 21 board-certificated dermatologists \cite{b5}. Brinker et al in 2019 for first time proved CNN model performance was superior to human assessment \cite{b21}. They trained ResNet50 on ISIC archive images with more than 20,000 dermatoscopy images yielded sensitivity and specitivity of 89.4\% and 68.2\% compare to that of 89.4\% and 64.4\% from 145 dermatologist on clinical images in MED-NODE database. With dermatoscopy images, a CNN can achieve higher results. A model based on MobileNet develop by Mohamed et al reached 92.7\% accuracy on the HAM10000 dataset \cite{b22}. Chaturvedi et al also used HAM10000 in \cite{b23} and deployed a fast-performance web application integrated with MobileNet model base that achieves 83.1\% accuracy. 
\section{Methodology}
\subsection{Dataset}
Although promising, early studies in skin cancer classification have been suffered from small dermatoscopy datasets. To bypass this problem, the HAM10000 dataset was published in 2018 with a great number of dermoscopy images [42]. The dataset is a collection of 10,015 skin lesion images of 7 different categories including (1) Actinic Keratosis, (2) Basal Cell Carcinoma, (3) Benign Keratosis, (4) Dermatofibroma, (5) Melanocytic nevi, (6) Melanoma, (7) Vascular Skin Lesion denoted as akiec, bcc, bkl, df, nv, mel, vasc, respectively. The images were acquired over the 20-year period from the Department of Dermatology at the Medical University of Vienna, Austria, and Skin cancer practice of Cliff Rosendahl Queensland, Australia. The images’ labels were confirmed either by histopathology, reflectance confocal microscopy, follow-up, or by expert consensus. 
Dataset images have 600 x 450-pixel resolution. Although the dataset consists of 10,015 skin lesion images, examination of the metadata indicates that there are only 7,470 distinct skin lesions. The remaining images are duplicates of different magnifications or different angles of view as shown in figure 1. Also, the dataset is extremely imbalanced, which is shown in  figure 2. For instance, the largest class (nv) contains 6,705 images while the smallest class, df, has only 115 images.

\begin{figure}[ht]
    \centering
    \includegraphics[width=0.7\linewidth, height=1.75cm]{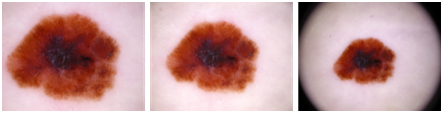} 
    \caption{Duplication with multiple angles}
    \label{fig:subim1}
\end{figure}

\begin{figure}[ht]
    \centering
    \includegraphics[width=0.9\linewidth, height=4.0cm]{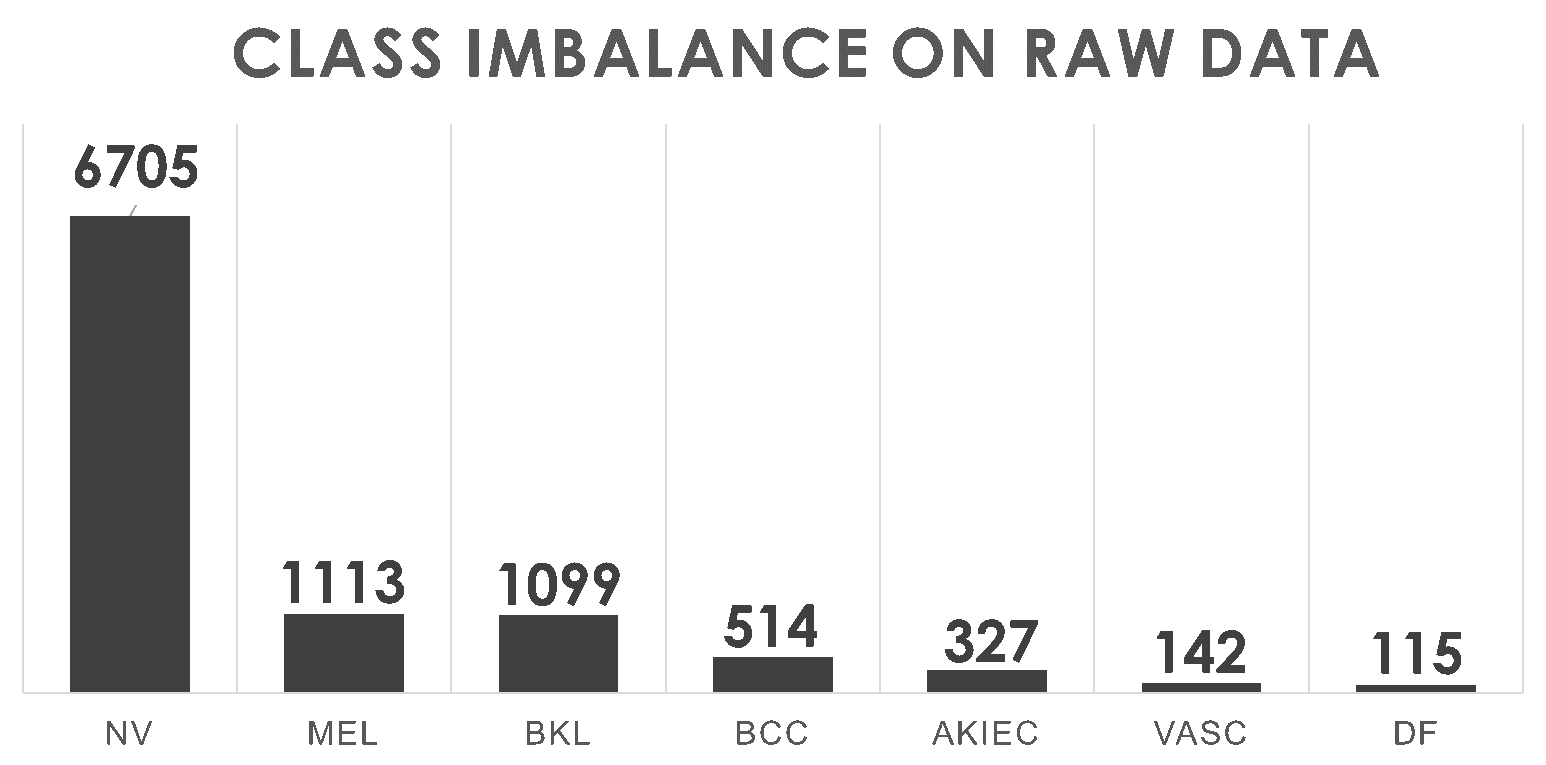}
    \caption{Imbalance in raw data}
    \label{fig:subim2}
\end{figure}

\subsection{Data preparation}
To train and test our AI models, images underwent a duplicate removal step. The result was a dataset without duplicates of 7,470 distinct lesion images. Subsequently, twenty percent of the non-duplicate dataset was held out for testing, called the test set (1,494 images). The remaining of the non-duplicate dataset was divided into the training set (6,817 images) and the validation set (1,704 images) with ratio of 80:20. Next, images in the training set were augmented by rotating, flipping, cut out and cropping operations. At the end of preprocessing, all samples were resized into 224x224 pixels. Stratified sampling was applied in every stage to maintain inter-class ratio among subsets and avoid the probability of missing out minority class between subsets. In the final experiment, to evaluate the consistency of final model over HAM10000, stratified 5-fold cross validation was applied in train-val split. The whole process is illustrated in figure 3.

\begin{figure*}[htp]
    \centering
    \includegraphics[width=\textwidth]{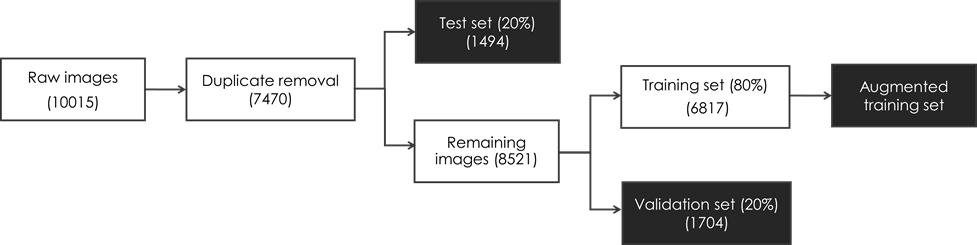}
    \caption{Train – validation - test set preparation process}
    \label{fig:fig3}
\end{figure*}

\begin{figure*}[htp]
    \centering
    \includegraphics[width=\textwidth]{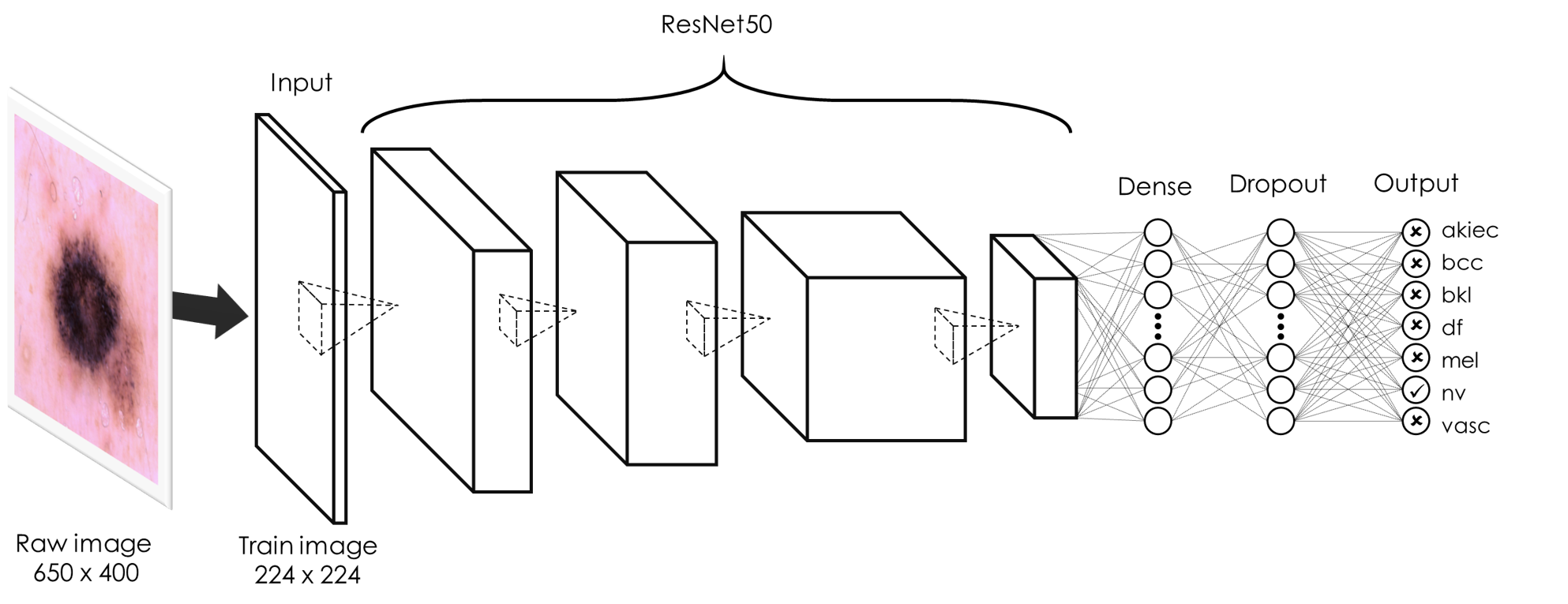}
    \caption{We used an improved ResNet50 model, tailored for the HAM10000 dataset, to classify these skin cancer images}
    \label{fig:fig}
\end{figure*}

\subsection{Model Architectures}
We applied transfer learning for skin lesion classification by slightly modifying architecture and fine-tuning weights of the ResNet50 \cite{b34} models pre-trained on the ImageNet dataset. The modification included: (1) use global average pooling instead of average pooling; (2) replace the top layer of the Resnet50 model with a block of a dropout layer (0.5) sandwiched by two fully connected layers (Fig. 4). The training was implemented using Google Colab with all layers of the pre-trained ResNet50 model unfrozen to learn new features of the HAM10000 dataset. Global average pooling and dropout layer were used to minimize overfitting by reducing some of the details of the learned features.

To address the class imbalance, we applied class-weighted learning approach by giving different weights for different classes in the loss function \cite{b35}. Higher weights were given to the minority classes whereas, lower weights are assigned to the majority classes. Weights were initialized based on ratios between the seven classes. These weights were later modified to increase accuracy with the final weights.

To further enhance the model performance, we employed a loss function called focal loss \cite{b36}. Focal loss is a special type of categorical loss that seeks a solution to data imbalance. The novel focal loss focuses on training on a sparse set of hard examples and prevents the vast number of easy negatives from overwhelming the detector during training. This idea is mathematically expressed as Eq. 1.
Where p denotes the predicted probability of the ground-truth class,  \(\alpha_t\) and \(\gamma\) are hyperparameters of the loss function.
\begin{equation} FL(p_t) = -\alpha_t * (1- p_t)^{\gamma} * \log(p_t)
\label{fc} \end{equation}
Obviously, with focal loss, easy classified samples are diminished while hard samples contribute with a larger amount to the loss values, which thereby make the model pay more attention to these samples and as a result, improves the precision for hard samples.

\section{Evaluation metrics}
These metrics are developed from True positive (TP), True negative (TN), False positive (FP) and False negative (FN) predictions.

\textbf{Accuracy:} 
reflect the amount of correct predictions (TP and TN) over all predictions (TP + FP + TN + FN). 
\begin{equation} Accuracy = \frac{TP+TN}{TP+FP+TN+FN}
\label{acc} \end{equation}
\textbf{Precision:} indicates true positive probability in all positive prediction cases. If the prediction is 1, all positive predictions are truly positive, however, there are still positive samples which are incorrectly predicted as negative.
\begin{equation} Precision = \frac{TP}{TP+FP}
\label{pre} \end{equation}
\textbf{Sensitivity: }or Recall is contrast to Precision as it indicates the probability of truly being negative when the prediction is negative. Similarly, if recall is 1, all negative predictions are truly negative, however, there are still negative samples which are incorrectly predicted as positive.
\begin{equation} Sensitivity = \frac{TP}{TP+FN}
\label{sen} \end{equation}
\textbf{F1-score: }established a balance between Precision and Recall. More advanced than accuracy, F1-score focuses on true positive value and is a better measurement for imbalance distribution classes.
\begin{equation} F1-score = \frac{2TP}{2TP+FP+FN}
\label{f1} \end{equation}
\textbf{Specificity:} indicate how well the model can detect negatives. Particularly this problem, due to the extremely imbalance dataset, sensitivities are usually high. 
\begin{equation} Specificity = \frac{TN}{FP+TN} 
\label{sepe} \end{equation}

\textbf{AUC - ROC curve:} is an overall performance measurement for classification problems at various thresholds settings. The ROC curve (or the receiver operating characteristic curve) is a probability curve and AUC (the Area under the curve) represents degree or measure of separability. It tells how much a model is capable of distinguishing between classes. Higher the AUC, better the model is at predicting 0s as. 0s and 1s as 1s.

\section{Results and Discussions}
In general, overfitting and class imbalance were addressed by five different techniques: drop out, augmentation, class-weighted (CW), focal loss (FC), global average pooling (GAP). To test the effectiveness of each method, 6 models were compiled, including 1 model that applies all techniques and 5 models that eliminate a single one. This idea can be summarized as Table I and the results of them are provided in Table II. For all experiments, learning rate reduce on plateau with initial learning 0.0001, Adam optimizer and checkpoint were utilized to train and save the optimal parameters for the model. The modified ResNet50 model which applied all methods achieved 88\% accuracy on validation set after 26 epochs with batch size of 64. Evaluating on the test set, the modified ResNet50 model gave result of 90\% accuracy. 
\renewcommand{\arraystretch}{1.25}
\vspace*{-5pt}
\begin{table}[ht]
\vspace*{-5pt}
    \centering
    \caption{Summary of models to test methods' effectiveness }
    \begin{tabular}{  c  c  c  c  c  c  } 
        \hline
Experiment&Dropout&	Augment&	CW&FC	&GAP\\
        \hline
1 (no dropout)	 &\textcolor{red}{\textbf{\textit{x}}} &\checkmark &\checkmark &\checkmark &\checkmark \\
2 (no augment) &\checkmark &\textcolor{red}{\textbf{\textit{x}}}&\checkmark &\checkmark &\checkmark \\
3 (no CW)	&\checkmark &\checkmark &\textcolor{red}{\textbf{\textit{x}}}&\checkmark &\checkmark \\
4 (no FC)	&\checkmark &\checkmark &\checkmark &\textcolor{red}{\textbf{\textit{x}}}&\checkmark \\
5 (no GAP)	&\checkmark &\checkmark &\checkmark&\checkmark &\textcolor{red}{\textbf{\textit{x}}}\\
6 (full)		&\checkmark &\checkmark &\checkmark&\checkmark &\checkmark \\
        \hline
    \end{tabular}
\end{table} 

The results in Table II implied that a combination of all methods in experiment 6 is the best. Since overfitting and class imbalance were addressed, it would be interesting to investigate detailed results for 7 individual classes from all experiments. Comparisons in precision, recall and F1-score were reported between experiments in Table III to V. The highest score and the lowest score in one class are highlighted in blue and red, respectively. 
\begin{table}[ht]
\vspace*{-10pt}
    \centering
    \caption{Results of all models to test methods' effectiveness}
    \begin{tabular}{  c  c  c  c  c  c  } 
        \hline
Experiment &Accuracy	&Precision&	Recall	&F1-score\\
        \hline
1 (no dropout)		&\textbf{90}&78&77&77\\
2 (no augment)	   &89&79&71&74\\
3 (no CW)			  &\textbf{90}&\textbf{81}&77&79\\
4 (no FC)              &74&70&\textbf{80}&72\\
5 (no GAP)	         &88&77&73&75\\
6 (full)	               &\textbf{90}&\textbf{81}&\textbf{80}&\textbf{80}\\
        \hline
    \end{tabular}
\end{table} 
\begin{table}[ht]
\vspace*{-5pt}
    \centering
    \caption{Breakdown into precision}
    \begin{tabular}{  c  c  c  c  c  c c c c  } 
        \hline
        \multicolumn{9}{c}{Precision} \\
        \hline
Exp	&akiec	&bcc	&bkl	&df	&mel	&nv	&vasc	&Average\\
         \hline
1	&0.72	&0.79	&0.79	&\textcolor{red}{0.59}	&\textcolor{blue}{0.73}	&0.95	&\textcolor{red}{0.86}	&0.78\\
2	&\textcolor{red}{0.55}&0.79	&0.77	&\textcolor{blue}{0.86}	&0.66	&\textcolor{red}{0.94}		&\textcolor{blue}{1.00}	&0.79\\
3	&\textcolor{blue}{0.78}	&0.81	&0.77	&0.69&	0.68	&0.95	&\textcolor{blue}{1.00}	&\textcolor{blue}{0.81}\\
4	&0.75	&\textcolor{red}{0.77}	&\textcolor{red}{0.48}	&0.68	&\textcolor{red}{0.33}&	\textcolor{blue}{0.99}&0.90&	\textcolor{red}{0.70}\\
5 &0.64	&\textcolor{blue}{0.82}	&0.75	&0.77	&0.62	&\textcolor{red}{0.94}	&\textcolor{red}{0.86}	&0.77\\
6 	&0.70	&\textcolor{blue}{0.82}	&\textcolor{blue}{0.84}&0.80&	0.72&\textcolor{red}{0.94}	&\textcolor{red}{0.86}	&\textcolor{blue}{0.81}\\
        \hline
    \end{tabular}
\end{table} 
\begin{table}[ht]
\vspace*{-18pt}
    \centering
    \caption{Breakdown into Sensitivity (Recall)}
    \begin{tabular}{  c  c  c  c  c  c c c c  }      
         \hline
        \multicolumn{9}{c}{Sensitivity (Recall)} \\
         \hline
Exp	&akiec	&bcc	&bkl	&df	&mel	&nv	&vasc	&Average\\
         \hline
1	&0.57	&0.77	&0.70	&0.67	&0.69	&\textcolor{blue}{0.97}	&\textcolor{blue}{1.00}	&0.77\\
2	&0.52	&0.74	&\textcolor{red}{0.67}	&\textcolor{red}{0.40}	&0.67	&\textcolor{blue}{0.97}	&\textcolor{blue}{1.00}	&\textcolor{red}{0.71}	\\
3	&\textcolor{blue}{0.70}	&0.78	&0.77	&0.60	&0.61	&\textcolor{blue}{0.97}	&\textcolor{blue}{1.00}	&0.77\\
4&	0.65&	\textcolor{red}{0.71}	&\textcolor{blue}{0.83}	&\textcolor{blue}{0.87}&\textcolor{blue}{0.85}&\textcolor{red}{0.72}	&\textcolor{red}{0.95}	&\textcolor{blue}{0.80}\\
5	&\textcolor{red}{0.46}	&0.77	&0.72	&0.67	&\textcolor{red}{0.54}	&\textcolor{blue}{0.97}	&\textcolor{blue}{1.00}	&0.73\\
6	&0.65&	\textcolor{blue}{0.83}	&0.70	&0.80	&0.63	&\textcolor{blue}{0.97}	&\textcolor{blue}{1.00}	&\textcolor{blue}{0.80}\\
        \hline
    \end{tabular}
\end{table} 
\begin{table}[ht]
\vspace*{-5pt}
    \centering
    \caption{Breakdown into F1 - score}
    \begin{tabular}{  c  c  c  c  c  c c c c  } 
        \hline
        \multicolumn{9}{c}{F1 - score} \\
        \hline
Exp	&akiec	&bcc	&bkl	&df	&mel	&nv	&vasc	&Average\\
        \hline
1	&0.63	&0.78&	0.74&	0.62	&\textcolor{blue}{0.71}&\textcolor{blue}{0.96}	&0.93&	0.77\\
2	&\textcolor{red}{0.53}&	0.76	&0.72&	\textcolor{red}{0.55}	&0.66	&\textcolor{blue}{0.96}&	\textcolor{blue}{1.00}	&0.74\\
3	&\textcolor{blue}{0.74}	&0.80	&\textcolor{blue}{0.77}&	0.64&	0.64	&\textcolor{blue}{0.96}&\textcolor{blue}{1.00}	&\textcolor{blue}{0.80}\\
4	&0.70	&\textcolor{red}{0.74}	&\textcolor{red}{0.61}&	0.76	&\textcolor{red}{0.48}&	\textcolor{red}{0.84}&\textcolor{red}{0.92}	&\textcolor{red}{0.72}\\
5&\textcolor{red}{0.53}	&0.79	&0.74	&0.71&0.58	&0.95	&0.93&	0.75\\
6	&0.67	&\textcolor{blue}{0.82}	&\textcolor{blue}{0.77}	&\textcolor{blue}{0.80}	&0.67	&\textcolor{blue}{0.96}&	0.93	&\textcolor{blue}{0.80}\\
        \hline
    \end{tabular}
\end{table} 

Besides the architecture as illustrated in Fig. 4, we also trained two other architectures by replacing the ResNet50 block in Fig. 4 by either VGG16, MobileNet or EfficientNetB1. Performance of the three architectures were compared using sensitivity, precision, and F1-score. The results in Table VI showed that ResNet50 model outperformed other models. However, in regard to deployment on mobile platforms, EfficientNetB1 could be a trade-off option due to its small size yet slightly lower accuracy than ResNet50’s accuracy.
\begin{table}[ht]
    \centering
    \caption{Results from stage 1}
    \begin{tabular}{  c  c  c  c  c  c } 
        \hline
Exp	&Methods	&Accuracy	&Precision&	Recall&	F1-score\\
        \hline
6	&ResNet50	&\textbf{90}&\textbf{81}&\textbf{80}&\textbf{80}\\
7	&VGG16	&85	&73	&61	&63\\
8	&MobileNet	&88	&76	&72	&73\\
9	&EfficientNet B1	&89 &76	&76	&75\\
        \hline
    \end{tabular}
\end{table} 

So far, ResNet50 has undergone many experiments and got top-notch tools to deal with HAM10000. However, it is essential to re-evaluate the model to examine its consistency, and robustness across the whole dataset.  Stratified 5-fold cross validation was used and 5 models were ensemble to get the best solution. Further evaluation of the ability of generalization was test by the ensemble model by test time augmentation (TTA) as well. Besides, for this final model,  top-2, top-3 accuracy and AUC-ROC score of all models were evaluated. All results are reported in Table VII, Fig.5 and Fig. 6. 
\begin{table}[ht]
    \centering
    \caption{Results from 5-folds cross validation}
    \begin{tabular}{  c  c  c  c  c  c  c c} 
        \hline
Fold&	Acc	&Pre	&Recall&	F1-score	&Top 2	&Top 3	&AUC\\
        \hline
1	&90	&85	&80	&82	&96.85	&98.73	&98.62\\
2	&90	&82	&77	&79	&96.18	&98.73	&98.39\\
3	&91	&83	&81	&82	&96.99	&99.00	&98.32\\
4	&91	&83	&82	&83	&96.85	&98.46	&98.48\\
5	&91	&85	&78	&81	&\textbf{97.46}	&\textbf{99.33}	&98.62\\
Ensemble	&\textbf{93}	&\textbf{88}	&84	&86	&\textbf{97.46}	&\textbf{99.33}	&\textbf{98.63}\\
By TTA	&\textbf{93}	&\textbf{88}	&\textbf{85}	&\textbf{87}	&96.85	&99.20	&98.41\\
        \hline
    \end{tabular}
\end{table} 
\begin{figure*}[htp]
    \centering
    \includegraphics[width=0.7\linewidth, height=6.0cm]{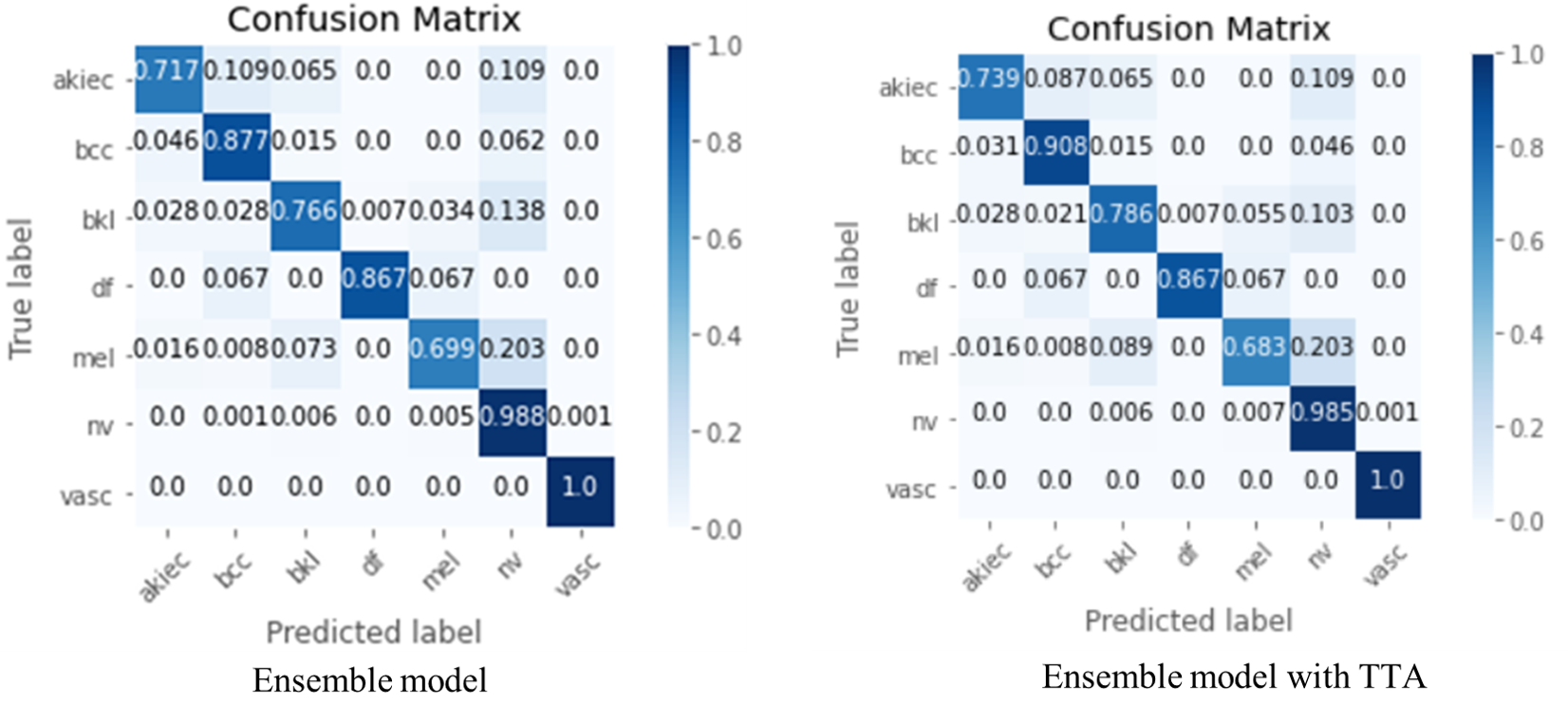}
    \caption{Confusion matrix of Ensemble model}
    \label{fig:fig4}
\end{figure*}
\begin{figure*}[htp]
    \centering
    \includegraphics[width=0.7\linewidth, height=6.0cm]{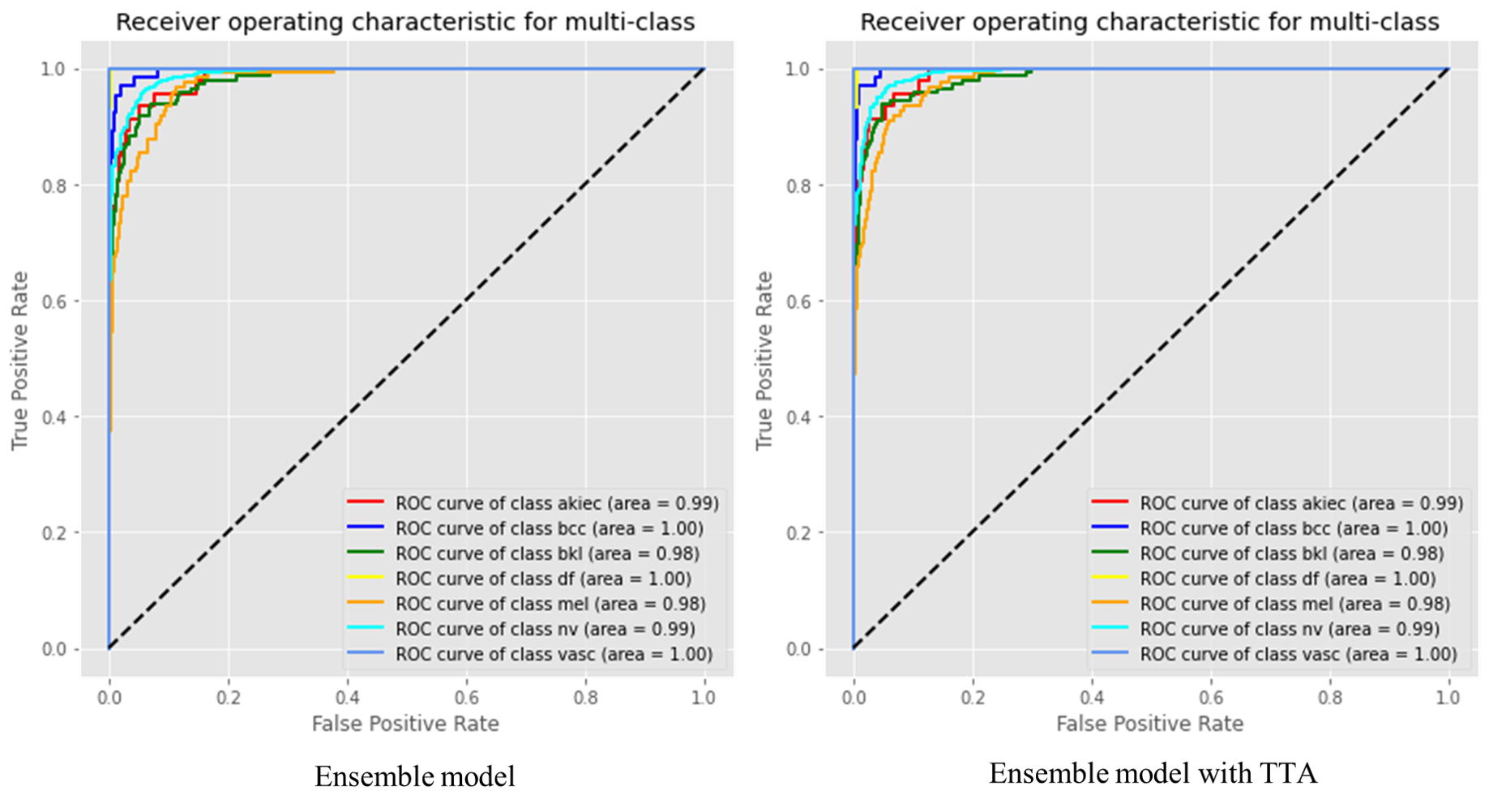}
    \caption{ROC plot of Ensemble model}
    \label{fig:fig5}
\end{figure*}
\subsection{Model interpretation:}
GradCAM was utilized to visualize and understand where the model looked in each sample for learning and predicting. The yellow area indicates the highest region where the model is focused on. In most cases, the model was looking for the right target – the lesion area in the image. However, there are some cases model confused by artifacts such as hair or bubbles. (Fig.7 and 8)
\begin{figure}[htp]
    \centering
    \includegraphics[width=1\linewidth, height=3.0cm]{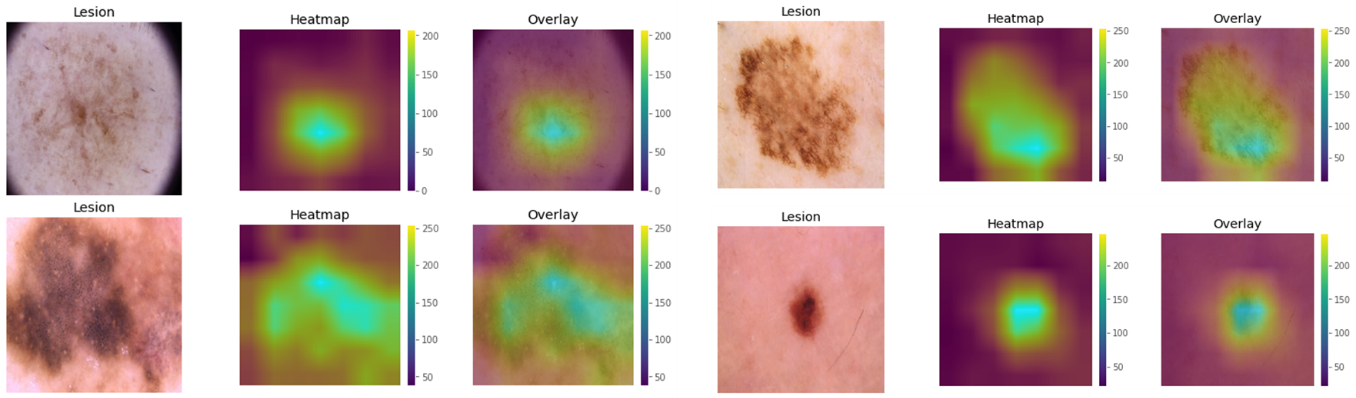}
    \caption{GradCAM visualization of good performance}
    \label{fig:fig10}
\end{figure}
\begin{figure}[ht]
    \centering
    \includegraphics[width=1\linewidth, height=3cm]{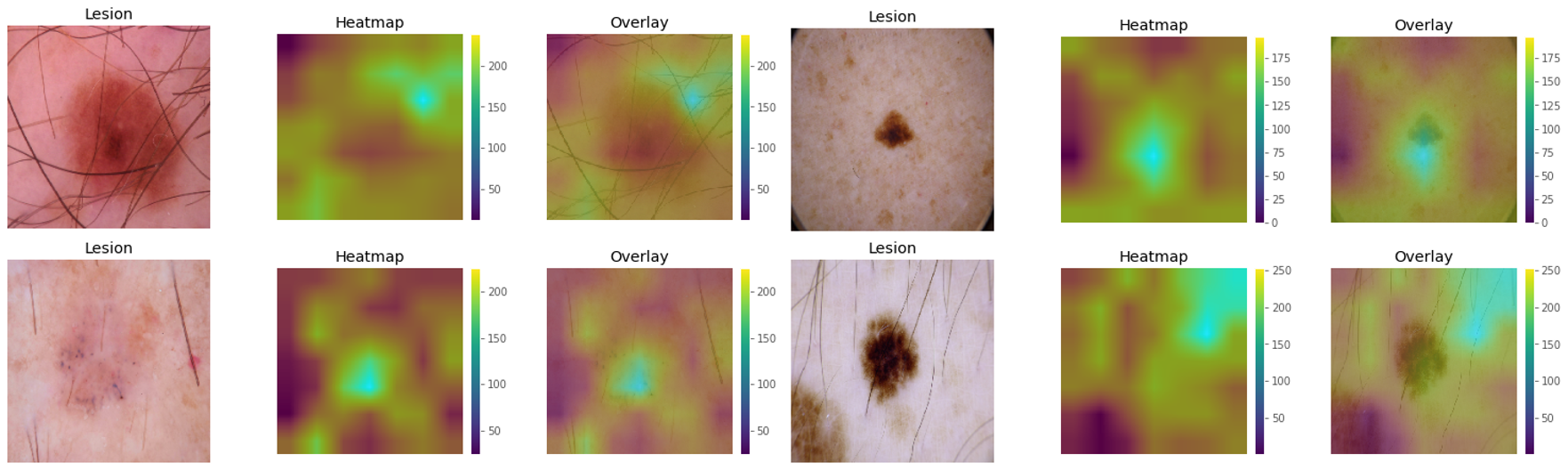}
    \caption{GradCAM visualization of poor performance}
    \label{fig:fig11}
\end{figure}
\subsection{Discussions:}
As observed from GradCAM, in some cases, model mistook areas outside lesions to diagnose. To tackle this problem, we built a simple segment model Unet with ResNet34 as backbone to segment lesions area as a new dataset to train model with different mask size by gradually increase dilating size of the mask. However, model performance significantly reduced. By experiments with different sizes of dilation, the larger the dilation, the higher the accuracy. Hence, we believe that model must also learn from areas surrounding lesions.

Overall, the performance of the whole system reached an accuracy of 93\% thanks to the aid of ensemble learning, F1-score is high indicating a balance between precision and recall. However, the individual model also has a high and acceptable performance, which can be effectively higher than performance of clinicians and be comparable against other works in Table VIII.
\vspace*{-5pt}
\begin{center}
\begin{table}[htbp]
\centering
\caption{Comparison to other works on HAM10000}
\begin{tabular}{  c  c  c  c  c  c c c  } 
     \hline
Work	&Year	&Model	&Acc&Pre	&Rec	&FS\\
     \hline
\cite{b24}&2020	&EfficientNetB1&	\textbf{94.00}&	94&	94	&94\\
\cite{b22}&2019	&MobileNet&	92.70&	87	&81&	84\\
\cite{b25}&2020	&Semi-supervised	&92.54	&	&71.47	&60.68\\
\cite{b26}&2020	&InceptionV3&	89.81&	88	&75&	81\\
\cite{b27}&2020	&ResNet50	&89.28		&&81	&81.28\\
\cite{b28}&2019	&Inception-ResNet	&83.96	&72.00	&69.29&69.86\\
\cite{b30}&2019	&MobileNet	&83.23		&&85	&82\\
\cite{b30}&2019	&VGG16	&82.80		&&64.57&	70\\
\cite{b31}&2019	&MobileNet	&82.00			&&&71\\
\cite{b32}&2020	&From scratch 	&80.93	&68.97	&53.95&58\\
\cite{b33}&2019	&From scratch 	&78.00		&&85	&86\\
\multirow{2}{1em}{\textbf{Ours}}&\multirow{2}{1em}{\textbf{2020}}	&\textbf{ResNet50}	&91	&83	&82&	83\\
		&&\textbf{Ensemble}	&\textbf{93.00}&	88	&86	&84\\
  \hline
\end{tabular}
\end{table}
\end{center}
\vspace*{-5pt}
In comparison, this proposed model is better than most of the studies and only behind the work in[24]. In [24], the authors also examined multiple models and came up with EfficientNetB1 with gains 94\% in 4 metrics. However, the problem is they only split the dataset into 2 subsets as training and validation. As they use validation set in training process for back propagation, the model will gradually adapt to weights that perform well in validation set. Without an extra test set as this study, it cannot reflect the true generalization and classification ability of the model on unseen data. Meanwhile, work in [22] ran into a similar problem when holding out only 552 images for testing. This amount of data is small compared with 8470 images in their dataset, thus, cannot fully indicate the robustness of the model. Furthermore, with identical base model ResNet50, the proposed method achieve better results than the study in [27]. In  [27], Mohammed A. Al-masni el investigated 2 stages diagnostic model comprised lesion boundaries segmentation and classification of those segmented lesions. Authors used ResNet50 as classifier by leveraging global average pooling after the last residual block and replace top layer with a fully connected with softmax for 7 classes classification. In attempt to confront class imbalance, weighted class and augmentation were applied. Indeed, this study is a good approach and their study assure that models will learn solely on lesion regions but surrounding skin regions. However, they might achieve better performance using some of the methods that we proposed in this work. 

\section{Conclusion}
In this study, we trained an end-to-end deep learning model without preprocessing steps or feature handcraft selection. We propose a modified ResNet50 deep learning model for classification of skin lesion images in the HAM10000 dataset. By reusing and modifying pre-trained model architecture together with training technique such as focal loss and class-weighted, our model achieved 93\% average accuracy and precision in the range [0.7, 0.94], which outperformed dermatologists’ accuracy of 84\%. Furthermore, we compared ResNet50 model against the VGG16, MobileNet and EfficientNet B1 as base model in our work. The results showed that ResNet50 model yielded the best performance. These recorded results could serve as a reference for future studies, helping scientists review training strategies and accelerate the training process. This work can be integrated in many complex systems for skin lesions analyzing. However, there are still some limitations in our work. For further improvement, we will try to remove artifacts that create biases for the model, expand the dataset with both dermoscopy images and clinical images, apply SOTA noisy student and pruning model after all to find the lottery ticket model to effectively reduce the size of the model.

\end{document}